\documentclass[11pt,a4paper]{article}
\usepackage[hyperref]{acl2019}
\usepackage[hyphenbreaks]{breakurl}

\usepackage{times}
\usepackage{latexsym}
\usepackage{soul}
\usepackage{multirow}
\usepackage{booktabs}
\usepackage{tabularx}
\usepackage{url}
\usepackage{todonotes}

\newcolumntype{s}{>{\hsize=.05\hsize}X}

\aclfinalcopy 
\def\aclpaperid{78} 

\title{The Unreasonable Effectiveness of Transformer Language Models in Grammatical Error Correction}

\author{Dimitris Alikaniotis \qquad Vipul Raheja\\
  Grammarly \\
  New York City, NY \\
  \texttt{firstname.lastname@grammarly.com} \\}

\date{}

\begin{document}
\maketitle
\begin{abstract}
  Recent work on Grammatical Error Correction (GEC) has highlighted the importance of language modeling in that it is certainly possible to achieve good performance by comparing the probabilities of the proposed edits. At the same time, advancements in language modeling have managed to generate linguistic output, which is almost indistinguishable from that of human-generated text. In this paper, we up the ante by exploring the potential of more sophisticated language models in GEC and offer some key insights on their strengths and weaknesses. We show that, in line with recent results in other NLP tasks, Transformer architectures achieve consistently high performance and provide a competitive baseline for future machine learning models.
\end{abstract}

\section{Introduction}

Transformer models \cite{DBLP:conf/nips/VaswaniSPUJGKP17} trained on large-scale language modeling datasets have recently proved to be a very effective means of representing the meaning of a sentence, being put to effective use in fine-tuning both  sentence-level tasks, such as the GLUE benchmark \cite{wang-etal-2018-glue} and token-level tasks, such as Named Entity Recognition \citep{DBLP:journals/corr/abs-1810-04805}. Recent work has also found them to produce linguistically valid representations \cite{DBLP:journals/corr/abs-1901-05287}, as well as to display excellent performance across multiple downstream NLP tasks  (e.g., \citealt{pmlr-v97-houlsby19a}).

In this work, we explore how such models perform in the task of Grammatical Error Correction (GEC). While there is a substantial amount of work on statistical \cite{rozovskaya-roth-2016-grammatical, junczys-dowmunt-grundkiewicz-2014-amu,yannakoudakis-etal-2017-neural} and neural \cite{ ji-etal-2017-nested, DBLP:journals/corr/XieAAJN16, yuan-briscoe-2016-grammatical, chollampatt2016ngec, chollampatt2017smtgec, sakaguchi-etal-2017-grammatical, chollampatt2018mlconv} machine translation methods for GEC, we follow the approach of  \citet{bryant-briscoe-2018-language} and explore how such models would fare in this task when treated as simple language models. More specifically, \citet{bryant-briscoe-2018-language} train a 5-gram language model on the One Billion Word Benchmark \citep{41880} dataset and find that it produces competitive baseline results without any supervised training. In our work, we extend this work by substituting the \textit{n}-gram model for several publicly available implementations of state-of-the-art Transformer language models trained on large linguistic corpora and assess their performance on GEC without any supervised training. We find that Transformer language models produce results on par with supervised approaches providing a solid baseline system. This finding is of particular importance in GEC, where data collection and annotation requires substantial manual effort.

\section{Related Work}
The idea of using language models is quite fundamental to the task of Grammatical Error Correction, which has fed a substantial body of work over the years. More recently, with the availability of web-scale data powering the advances in language modeling, among most of the other advances in NLP, a plethora of language-modeling based approaches have been proposed for the GEC task. \citet{gamon-etal-2008-using, HERMET08.220} and \citet{yi-etal-2008-web} were some   of the early works to successfully leverage language models trained on large amounts of web-scale data into a GEC system, reinforcing the idea that simple models and a lot of data trump more elaborate models based on annotated data \cite{4804817}. 

Since then, multiple works based on language-models have been proposed for the GEC task \cite{Park:2011:AWS:2002472.2002590, dahlmeier-ng-2012-beam}, either relying entirely on LMs or using them for fine-tuning their systems. Many of the top-ranked systems in the CoNLL-2013 and 2014 GEC shared tasks \cite{ng-etal-2013-conll,ng-etal-2014-conll}, were either based on language models or had them as integral parts of their systems \cite{kao-etal-2013-conll, yoshimoto-etal-2013-naist, xing-etal-2013-um,lee-lee-2014-postech, junczys-dowmunt-grundkiewicz-2014-amu}. LM-only approaches though took a backseat and were only sporadically used after the shared tasks, as Neural Machine Translation-based approaches took over, but LMs remained an integral part of the GEC systems \cite{junczys-dowmunt-grundkiewicz-2016-phrase,ji-etal-2017-nested, DBLP:journals/corr/XieAAJN16, junczys-dowmunt-etal-2018-approaching, chollampatt2018mlconv}. However, \citet{bryant-briscoe-2018-language} recently revived the idea, achieving competitive performance with the state-of-the-art, demonstrating the effectiveness of the approaches to the task without using any annotated data for training. 

\section{Methodology}

In this work, we follow the setup from \citet{bryant-briscoe-2018-language} substituting the 5-gram language model for different language models based on the Transformer architecture. Specifically, we use Google's BERT \citep{DBLP:journals/corr/abs-1810-04805} and OpenAI's GPT \citep{radford2018improving} and GPT-2 \citep{noauthororeditor}.\footnote{\url{https://github.com/huggingface/pytorch-pretrained-BERT/}} While all these are best thought of as language models in that they have been trained to predict an element in a sequence, they use slightly different objectives which does not make them directly comparable. Specifically, GPT and GPT-2 have been trained with a classic language modeling objective, whereby they predict the next word in a sequence, whereas BERT has been trained using a \textit{masked language modeling} objective in which the network attempts to predict masked words in the sentence.

We extract the probability of a sentence from BERT, by iteratively masking every word in the sentence and then summing the log probabilities. While this approach is far from ideal, it has been shown \citep{DBLP:journals/corr/abs-1902-04094} that it approximates the log-likelihood of a sentence.

\begin{table}[t]
    \centering
    \begin{tabular}{cccc}
        \toprule
        Test set & Sent. & Tokens & Annot. \\ \midrule
        CoNLL-2014 & 1,312 & 30k & 2 \\
        FCE & 2,715 & 47k & 1 \\\bottomrule
    \end{tabular}
    \caption{Statistics for evaluation data}
    \label{tab:data}
\end{table}

\subsection{Confusion sets}

Since our systems do not generate novel sequences, we follow \citet{bryant-briscoe-2018-language} and use simple heuristics to generate a confusion set of sentences that our language models score. For prepositions and determiners, the confusion set includes the set of all prepositions and determiners plus an empty string $\epsilon$ to remove unnecessary additions. For morphological errors (e.g., past tense or pluralization), we use the Automatically Generated Inflection Database (AGID) which contains different morphological forms for each word.\footnote{\url{http://wordlist.aspell.net/other/}} However, we notice that due to the automatic generation, AGID contains errors that might propagate into our scoring. The problem with introducing new errors and non-words is that they would be interpreted as unknown words (henceforth \texttt{[UNK]}s) from the model's perspective. An unknown word in some context might give higher probabilities to an erroneous sentence and cause the model not to select the correct alternative. To remedy this issue, we generate a vocabulary from all the training sets and make sure that any proposed words which do not exist in the vocabulary are replaced by \texttt{[UNK]}s. Note that there is no reason to re-use the vocabulary of the training sets as any large English wordlist would achieve a similar effect. Finally, for spelling mistakes, we, again, follow \citet{bryant-briscoe-2018-language} and use CyHunSpell\footnote{\url{https://pypi.org/project/CyHunspell/}} to generate alternatives for non-words.

\subsection{Thresholding}

\begin{table*}[t!]
\small
\begin{tabularx}{.99\linewidth}{lXlllslll}\toprule
 & & \multicolumn{3}{c}{\textbf{ERRANT}} &&  \multicolumn{3}{c}{\textbf{M2}}\\ \midrule
\textbf{Dataset} & \textbf{System} & \textbf{P} & \textbf{R} & \textbf{F$_{0.5}$} && \textbf{P} & \textbf{R} & \textbf{F$_{0.5}$}\\ \midrule
\multirow{9}{*}{\textbf{CoNLL-2014}}
& \citet{felice-etal-2014-grammatical} $\dagger$ & - & - & - & & 39.71 & 30.10 & 37.33 \\
& \citet{yannakoudakis-etal-2017-neural} & - & - & - & & 58.79 & 30.63 & 49.66\\
& \citet{chollampatt2017smtgec} & - & - & - & & 62.74 & 32.96 & 53.14\\
& \citet{chollampatt2018mlconv} & - & - & - & & 65.49 & 33.14 & 54.79\\
& \citet{DBLP:journals/corr/abs-1807-01270} & - & - & - & & \textbf{74.12} & \textbf{36.30} & \textbf{61.34}\\ \cmidrule{2-9}
 & \citet{bryant-briscoe-2018-language} & 36.62 & 19.93 &  31.37 && 40.56 & 20.81 & 34.09 \\
& BERT  & 33.27	&\textbf{27.14}&	31.83 && 35.69 & \textbf{27.99}          & 33.83 \\ 
& GPT-1 & 49.58&	27.06&	42.5 && 51.08 & 27.45                   & 43.57 \\
& GPT-2 & \textbf{57.73}	& 24.75 & \textbf{45.58} && \textbf{58.51} & 24.9 & \textbf{46.08} \\\midrule\midrule

\multirow{5}{*}{\textbf{FCE}} 
& \citet{yannakoudakis-etal-2017-neural} & - & - & - && \textbf{65.03} & 32.45 & \textbf{54.15} \\ \cmidrule{2-9}
& \citet{bryant-briscoe-2018-language} & 41.92 & 13.62 & 29.61 && 44.78 & 14.12 & 31.22 \\
& BERT  & 29.56 & \textbf{34.67}  & 30.46 && 31.97 & \textbf{35.01} & 32.53  \\
& GPT-1 & \textbf{62.75} & 32.19 & 52.74 && \textbf{64.01}  & 32.33 & \textbf{53.52} \\
& GPT-2 & 61.91  & 33.47 & \textbf{52.92} && 62.64 & 33.74  & 53.48  \\\midrule

\end{tabularx}
\caption{Results of our Transformer-Language Model approach against similar approaches \citep{bryant-briscoe-2018-language} and state-of-the-art on Grammatical Error Correction. For each of the datasets, we use the corresponding test set, and we do not train our models on the corpora. As BERT, we report the best performing BERT model (12 layers, retaining uppercase characters). In the top part of each dataset, we report the scores of supervised methods and in the bottom the unsupervised ones. $\dagger$ denotes this system won the shared task competition.}
\label{data-table}
\end{table*}

Given that our confusion set is prone to errors (due to its automatic generation procedure) as well as the fact that we cannot target all potential errors (e.g., insertions), we bias our method to \textit{prefer} the original sentence unless a much better the alternative is found. We quantify this margin by imposing a threshold above which we accept a candidate sentence as a better alternative. Concretely, let $P(s_c)$ be the probability of the candidate sentence and $P(s_o)$ the probability of the original sentence, then we accept the candidate if $P(s_c) > P(s_o) + \tau$, where $\tau$ is some threshold parameter which we fit on each development set. Note that, practically, this parameter controls the trade-off between \textit{precision} and \textit{recall} as higher $\tau$ values would mean that there is less chance of changing the original sentence (i.e., higher precision) and vice versa. We explore different values for $\tau \in \{0, 2, 4, 6, 8\}$ by, as above, fitting them on the corresponding development set.\footnote{Note that the probability of each sentence is in log space.}

\subsection{Search}

Finally, we perform \textit{greedy search} to find the best alternative sentence by iterating over each sentence multiple times, once for every position for which our heuristics found alternatives. If an alternative is selected for the target position, we update the original sentence and proceed to the next position. This pseudo-log-likelihood approximation makes the problem of considering every permutation more computationally tractable.

\section{Experiments}

We evaluate our method and report results on two standard publicly available datasets. Our evaluation is aimed to stay as true to \citet{bryant-briscoe-2018-language} as possible to ensure an even comparison. Concretely, we use the test dataset from the CoNLL-2014 \cite{ng-etal-2014-conll} shared task\footnote{While we acknowledge the contemporaneous nature of the BEA 2019 Shared Task on GEC and would have liked to report results on the W\&I+LOCNESS data, we could not do so because of license limitations.} and the publicly available First Certificate in English (FCE) \cite{yannakoudakis-etal-2011-new}. Unfortunately, due to licensing issues, we were unable to obtain permission to use the JFLEG \cite{napoles-sakaguchi-tetreault:2017:EACLshort} corpus for evaluation. Note that in our method, we do not make use of the training sets commonly used with these datasets. However, we use the development sets used by \citet{bryant-briscoe-2018-language} to tune the hyperparameter $\tau$. The number of sentences and tokens for the datasets we used can be found in Table~\ref{tab:data}.

Similar to \citet{bryant-briscoe-2018-language}, we report results on three metrics. We use the MaxMatch (M$^2$) Precision, Recall and F$_{0.5}$ \cite{dahlmeier-ng-2012-better} and ERRANT Precision, Recall and F$_{0.5}$ \cite{bryant-etal-2017-automatic}.

\section{Results}

\begin{table*}[t!]

\begin{tabularx}{.99\textwidth}{lX}\toprule
\textbf{Source} & It will start by a speech from the Director of the conference, followed by a meal. \\
\textbf{Gold} & It will start \textbf{with} a speech \textbf{by} the Director of the conference, followed by a meal. \\
\textbf{BERT} & It will start \textbf{with} a speech \textbf{from} the Director of the conference, followed by a meal. \\
\textbf{GPT} & It will start \textbf{by} a speech \textbf{from} the Director of the conference, followed by a meal. \\
\textbf{GPT-2} & It will start \textbf{with} a speech \textbf{from} the Director of the conference, followed by a meal. \\\midrule
\textbf{Source} & They all knows where the conference is and when. \\
\textbf{Gold} & They all \textbf{know} where the conference is and when. \\
\textbf{BERT} & They all know where the \textbf{conferencing} is and when. \\
\textbf{GPT} & They all \textbf{knows} where the conference is and when. \\
\textbf{GPT-2} & They all \textbf{know} where the conference is and when. \\
\bottomrule
\end{tabularx}
\caption{Source sentences along with the gold edits and the proposed candidates from each of our models.}
\label{data-table-detection}
\end{table*}

Table \ref{data-table} presents the results of our method comparing them against recent state-of-the-art supervised models and the simple \textit{n}-gram language model used by \citet{bryant-briscoe-2018-language}. Table~\ref{data-table-detection} shows some qualitative examples on how each model corrects two sentences pulled from the FCE along with the gold annotations. The reported results come from the best performing hyperparameter $\tau$ on each dataset. For BERT, we also explored different sizes (12 vs. 24 layers) and whether retaining uppercase characters helps in performance. The best performing $\tau$ values were $\tau=4$ for CoNLL14 for all models; for the FCE dataset: BERT $\tau=4$, GPT $\tau=8$, and GPT-2 $\tau=6$. The best `version,' of BERT was the \textit{large}, \textit{cased} (i.e., retaining the lower- /uppercase distinction).

A key result of Table \ref{data-table} is that Transformer Language Models prove to be more than just a competitive baseline to legitimate Grammatical Error Correction systems on their own. Across the board, Transformer Models are able to outperform the simple \textit{n}-gram model and even approach the performance of supervised GEC systems.

\section{Discussion}
Looking at the performance of the two GPT models more closely, we see that their performance is nearly identical with GPT-2 leading by a small margin in the CoNLL14 dataset. Given that the  versions we used share the same number of layers (12), we attribute GPT-2's slight advantage to the fact that it was trained on considerably more data.

Another interesting result is that while BERT surpasses the \textit{n}-gram baseline overall, it achieves worse performance than the rest in terms of precision and F$_{0.5}$ score. Considering its overall success at modeling NLP tasks, one might expect BERT to achieve better performance here. However, as  mentioned above, BERT is not truly a language model in the sense that GPT and GPT-2 are but uses a quasi-language modeling objective which could explain its degraded performance in this setting. Note that framing the task differently (e.g., by masking the preposition in a sentence and selecting the one with the highest probability) might give the edge to BERT as it resembles the way it was trained.

It is also worth mentioning that despite tuning $\tau$ to each dataset, we do not explore different weights for different kinds of errors (e.g., penalizing more spelling mistakes). Our key motivation was to corroborate and extend the results of \citet{bryant-briscoe-2018-language} to current state-of-the-art language models which have been trained in several languages and show that these models are tough baselines to beat for novel GEC systems.

While the results of the Transformer language models shown in Table~\ref{data-table} demonstrate that they are a tough baseline to beat, it is worth noting that the present approach is not without its limitations.
We believe that our methodology should not be considered a panacea to GEC. For instance, being bound by the confusion sets, our system (1) cannot handle missing words (which make up about 20\% of all errors), and (2) it is tuned to capture only a subset of the possible mistakes a writer can make (closed class words).

It could be argued that since our system makes use of a pre-defined confusion set (even an automatically generated one), it could not be considered as a fully unsupervised system. In principle, we agree with that statement and we believe that a system which uses, for example, corpus statistics to on-the-fly generate a confusion set would be a very interesting exercise and could yield similar results. However, the present paper is concerned with highlighting the importance of language modeling in GEC and its potential in aiding in low-resource languages where large parallel datasets are unavailable, but such confusion sets are relatively easily obtainable.

\section{Conclusion}
In this work, we advanced on the foundational idea that a simple language modeling-based approach to GEC with no annotated data can challenge the latest neural and machine translation approaches that rely on large quantities of annotated training data. To this end, we improve on previous work by leveraging state-of-the-art language modeling techniques and perform a thorough comparison of three state-of-the-art Transformer language models which in turn have been trained on data of the order of hundreds of millions of words. We find that merely using pre-trained, and publicly available neural language models improves the performance by a significant margin and comes within striking distance of the state-of-the-art methods.

This work reinforces the strength and robustness of language-model based methods for the task of grammatical error correction. While recent state-of-the-art GEC systems are pursuing NMT-based models with huge amounts (millions of sentences) of annotated training data, approaches like this which require no annotated training data provide great value to researchers and developers interested in building competitive GEC systems (e.g., in other languages) with limited annotated data. 

\section*{Acknowledgements}

The authors would like to thank Joel Tetreault for his comments on earlier drafts as well as the anonymous reviewers for their helpful suggestions.

\bibliography{acl2019}
\bibliographystyle{acl_natbib}

\end{document}